\documentclass[11pt,letterpaper]{article}
\usepackage{emnlp2017}
\usepackage{times}
\usepackage{amsfonts}
\usepackage{amsmath}
\usepackage{latexsym}
\usepackage{graphicx}
\usepackage{tikz}
\usepackage{multirow}
\usepackage{subfigure}

\emnlpfinalcopy

\def\emnlppaperid{506}

\title{Globally Normalized Reader}

\author{Jonathan Raiman \and John Miller \\
Baidu Silicon Valley Artificial Intelligence Lab\\
  {\tt \{jonathanraiman,millerjohn\}@baidu.com}
}

\date{}

\begin{document}

\maketitle

\begin{abstract}
Rapid progress has been made towards question answering (QA) systems that can extract answers from text. Existing neural approaches make use of expensive bi-directional attention mechanisms or score all possible answer spans, limiting scalability. We propose instead to cast extractive QA as an iterative search problem: select the answer's sentence, start word, and end word. This representation reduces the space of each search step and allows computation to be conditionally allocated to promising search paths. We show that globally normalizing the decision process and back-propagating through beam search makes this representation viable and learning efficient. We empirically demonstrate the benefits of this approach using our model, Globally Normalized Reader (GNR), which achieves the second highest single model performance on the Stanford Question Answering Dataset (68.4 EM, 76.21 F1 dev) and is 24.7x faster than bi-attention-flow. We also introduce a data-augmentation method to produce semantically valid examples by aligning named entities to a knowledge base and swapping them with new entities of the same type. This method  improves the performance of all models considered in this work and is of independent interest for a variety of NLP tasks.

\end{abstract}

\section{Introduction}

\begin{figure}[ht]
    \centering
    \includegraphics[width=3in]{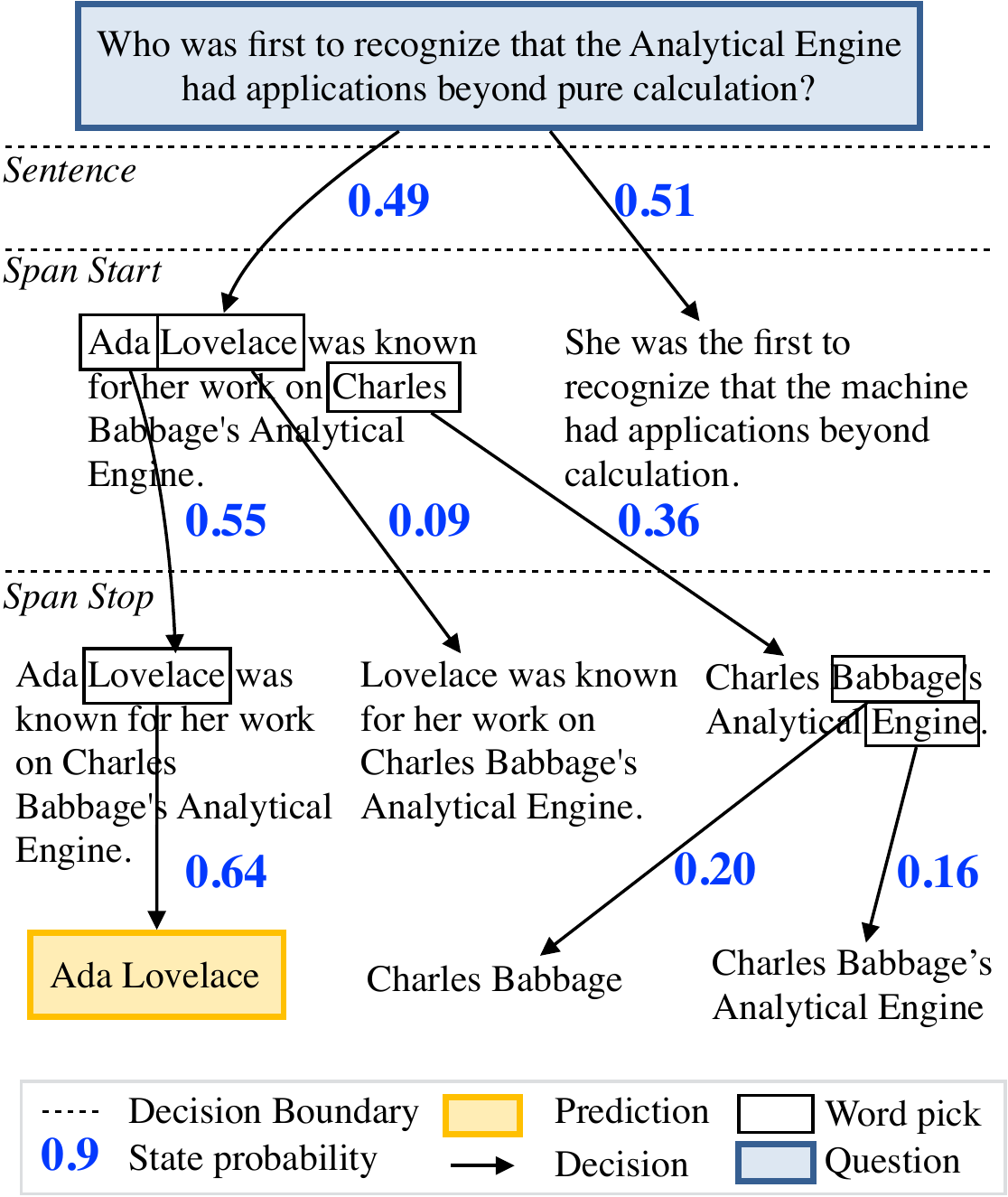}
    \caption{GNR answering a question. It first picks a sentence, then start word, then end word. Probabilities are global and normalized over the beam. Model initially picks the wrong sentence, but global normalization lets it recover. Final prediction's probability (0.64) exceeds sentence pick (0.49), whereas with local normalization each probability is upper bounded by the previous step.}
    \label{fig:example}
\end{figure}

Question answering (QA) and information extraction systems  have proven to be invaluable in wide variety of applications such as medical information collection on drugs and genes \cite{quirk2016distant}, large scale health impact studies \cite{althoff2016influence}, or educational material development \cite{koedinger2015data}. Recent progress in neural-network based extractive question answering models are quickly closing the gap with human performance on several benchmark QA tasks such as SQuAD \cite{rajpurkar2016squad}, MS MARCO \cite{nguyen2016ms}, or NewsQA \cite{trischler2016newsqa}. However, current approaches to extractive question answering face several limitations:
\begin{enumerate}
    \item Computation is allocated equally to the entire document, regardless of answer location, with no ability to ignore or focus computation on specific parts. This limits applicability to longer documents.
    \item They rely extensively on expensive bi-directional attention mechanisms \cite{seo2016bidirectional} or must rank all possible answer spans \cite{lee2016learning}.
    \item While data-augmentation for question answering have been proposed \cite{zhou2017neural}, current approaches still do not provide training data that can {\em improve} the performance of existing systems. 
\end{enumerate}

In this paper we demonstrate a methodology for addressing these three limitations, and make the following claims:

\begin{enumerate}
    \item Extractive Question Answering can be cast as a nested search process, where sentences provide a powerful document decomposition and an easy to learn search step. This factorization enables conditional computation to be allocated to sentences and spans likely to contain the right answer.
    \item When cast as a search process, models without bi-directional attention mechanisms and without ranking all possible answer spans can achieve near state of the art results on extractive question answering.
    \item Preserving narrative structure and explicitly incorporating type and question information into synthetic data generation is key to generating examples that actually improve the performance of question answering systems.
\end{enumerate}

Our claims are supported by experiments on the SQuAD dataset where we show that the Globally Normalized Reader (GNR), a model that performs an iterative search process through a document (shown visually in Figure \ref{fig:example}), and has computation conditionally allocated based on the search process, achieves near state of the art Exact Match (EM) and F1 scores without resorting to more expensive attention or ranking of all possible spans. Furthermore, we demonstrate that Type Swaps, a type-aware data augmentation strategy that aligns named entities with a knowledge base and swaps them out for new entities that share the same type, improves the performance of all models on extractive question answering.

We structure the paper as follows: in Section 2 we introduce the task and our model. Section 3 describes our data-augmentation strategy. Section 4 introduces our experiments and results. In Section 5 we discuss our findings. In Section 6 we relate our work to existing approaches. Conclusions and directions for future work are given in Section 7.

\section{Model}
\begin{figure*}[ht]
    \centering
    \includegraphics[width=\textwidth]{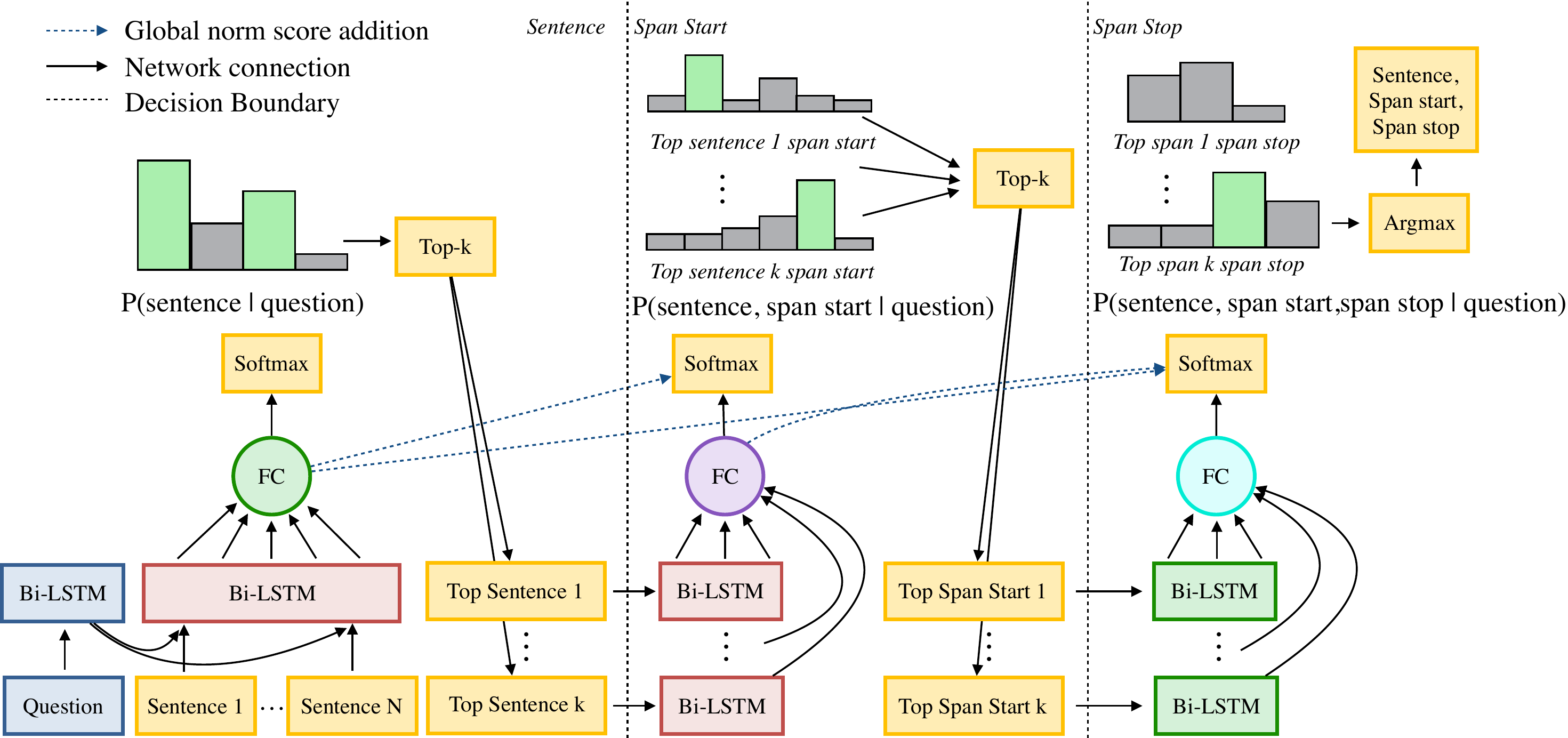}
    \caption{Globally Normalized Reader's search process. Same color Bi-LSTMs share weights.}
    \label{fig:network-sketch}
\end{figure*}

Given a document $d$ and a question $q$, we pose extractive question answering as a search problem. First, we select the sentence, the first word of the span, and finally the last word of the span. A example of the output of the model is shown in Figure \ref{fig:example}, and the network architecture is depicted in Figure \ref{fig:network-sketch}.

More formally, let $d_1, \dots, d_n$ denote each sentence in the document, and for each sentence $d_i$, let $d_{i,1}, \dots, d_{i,{m_i}}$ denote the word vectors corresponding to the words in the sentence. Similarly, let $q_1, \dots, q_\ell$ denote the word vectors corresponding to words in the question. An answer is a tuple $a = (i^*, j^*, k^*)$ indicating the correct sentence $i^*$, start word in the sentence $j^*$ and end word in the sentence $k^*$. Let $\mathcal{A}(d)$ denote the set of valid answer tuples for document $d$. We now describe each stage of the model in turn.

\subsection{Question Encoding}
Each question is encoded by running a stack of bidirectional LSTM (Bi-LSTM) over each word in the question, producing hidden states $(h_{1}^{\mathrm{fwd}}, h_{1}^{\mathrm{bwd}}), \dots, (h_{\ell}^{\mathrm{fwd}}, h_{\ell}^{\mathrm{bwd}})$ \cite{graves2005framewise}. Following \citet{lee2016learning}, these hidden states are used to compute a {\em passage-independent question embedding}, $q^{\text{indep}}$.  Formally,
\begin{align}
    s_j &= w_{q}^\top \mbox{MLP}([h_j^{\mathrm{bwd}}; h_j^{\mathrm{fwd}}]) \\
    \alpha_j &= \frac{\exp(s_j)}{\sum_{j'=1}^\ell \exp(s_{j'})} \\
    q^{\text{indep}} &= \sum_{j=1}^\ell \alpha_j [h_j^{\mathrm{bwd}}; h_j^{\mathrm{fwd}}],
\end{align}
where $w_q$ is a trainable embedding vector, and 
$\mbox{MLP}$ is a two-layer neural network with a $\mathrm{Relu}$ non-linearity. The question is represented by concatenating the final hidden states of the forward and backward LSTMs and the passage-independent embedding, $q = [h_1^{\mathrm{bwd}}; h_\ell^{\mathrm{fwd}};  q^{\text{indep}}]$.

\subsection{Question-Aware Document Encoding}
Conditioned on the question vector, we compute a representation of each document word that is sensitive to both the surrounding context and the question. Specifically, each word in the document is represented as the concatenation of its word vector $d_{i,j}$, the question vector $q$, boolean features indicating if a word appears in the question or is repeated, and a {\em question-aligned embedding} from \citet{lee2016learning}. The question-aligned embedding $q^{\text{align}}_{i, j}$ is given by
\begin{align}
    s_{i, j, k} &= \mbox{MLP}(d_{i, j})^\top \mbox{MLP}(q_k) \\
    \alpha_{i, j, k} &= \frac{\exp(s_{i, j, k})}{\sum_{k'=1}^{\ell} \exp(s_{i, j, k'})} \\
    q^{\text{align}}_{i, j} &= \sum_{k=1}^{\ell} \alpha_{i, j, k} q_k.
\end{align}
The document is encoded by a separate stack of Bi-LSTMs, producing a sequence of hidden states $(h_{1,1}^{\mathrm{fwd}}, h_{1,1}^{\mathrm{bwd}}), \dots, (h_{n,m_n}^{\mathrm{fwd}}, h_{n,m_n}^{\mathrm{bwd}})$. The search procedure then operates on these hidden states.

\subsection{Answer Selection}
\paragraph{Sentence selection.}
The first phase of our search process picks the sentence that contains the answer span. Each sentence $d_i$ is represented by the hidden state of the first and last word in the sentence for the backward and forward LSTM respectively,  $[h_{i,1}^{\mathrm{bwd}}; h_{i,{m_i}}^{\mathrm{fwd}}]$, and is scored by passing this representation through a fully connected layer that outputs the unnormalized sentence score for sentence $d_i$, denoted $\phi_\mathrm{sent}(d_i)$.

\paragraph{Span start selection.}
After selecting a sentence $d_i$, we pick the start of the answer span within the sentence. Each potential start word $d_{i,j}$ is represented as its corresponding document encoding $[h_{i,j}^{\mathrm{fwd}};h_{i,j}^{\mathrm{bwd}}]$, and is scored by passing this encoding through a fully connected layer that outputs the unnormalized start word score for word $j$ in sentence $i$, denoted $\phi_\mathrm{sw}(d_{i,j})$.

\paragraph{Span end selection.}
Conditioned on sentence $d_i$ and starting word $d_{i,j}$, we select the end word from the remaining words in the sentence $d_{i,j}, \dots, d_{i,m_i}$. To do this, we run a Bi-LSTM over the remaining document hidden states $(h_{i,j}^{\mathrm{fwd}}, h_{i,j}^{\mathrm{bwd}}), \dots, (h_{i,m_i}^{\mathrm{fwd}}, h_{i,m_i}^{\mathrm{bwd}})$ to produce representations $(\tilde{h}_{i,j}^{\mathrm{fwd}}, \tilde{h}_{i,j}^{\mathrm{bwd}}), \dots, (\tilde{h}_{i,m_i}^{\mathrm{fwd}}, \tilde{h}_{i,m_i}^{\mathrm{bwd}})$. Each end word $d_{i, k}$ is then scored by passing $[\tilde{h}_{i,k}^{\mathrm{fwd}}; \tilde{h}_{i,k}^{\mathrm{bwd}}]$ through a fully connected layer that outputs the unnormalized end word score for word $k$ in sentence $i$, with start word $j$, denoted $\phi_\mathrm{ew}(d_{i,j:k})$.

\subsection{Global Normalization}
The scores for each stage of our model can be normalized at the local or global level. Previous work  demonstrated that locally-normalized models have a weak ability to correct mistakes made in previous decisions, while globally normalized models are strictly more expressive than locally normalized models  \cite{andor2016globally, zhou2015neural, collins2004incremental}.

In a locally normalized model each decision is made conditional on the previous decision. The probability of some answer $a = (i, j, k)$ is decomposed as
\begin{equation}
\begin{split}
    \mathbb{P}(a|d,q) =& \mathbb{P}_{\mathrm{sent}}(i|d,q) \cdot\mathbb{ P}_{\mathrm{sw}}(j|i,d,q) \cdot\\& \mathbb{ P}_{\mathrm{ew}}(k|j, i,d,q).
\end{split}
\end{equation}
Each sub-decision is locally normalized by applying a softmax to the relevant selection scores:
\begin{equation}
\mathbb P_{\mathrm{sent}}(i|d,q) = \frac{
   \exp(\phi_\mathrm{sent}(d_i))
}{
   \sum_{x=1}^n \exp(\phi_\mathrm{sent}(d_x))
},
\end{equation}
\vspace{-1.1em}
\begin{equation}
\mathbb{P}_{\mathrm{sw}}(j|i,d,q) = \frac{\exp(\phi_\mathrm{sw}(d_{i,j}))}{\sum_{x=1}^{m_i} \exp(\phi_\mathrm{sw}(d_{i,x}))},
\end{equation}
\vspace{-1.1em}
\begin{equation}
\mathbb P_{\mathrm{ew}}(k|j,i,d,q) = \frac{\exp(\phi_\mathrm{ew}(d_{i,j:k}))}{
 \sum_{x=j}^{m_i} \exp(\phi_\mathrm{ew}(d_{i,j:x}))}.
\end{equation}

To allow our model to recover from incorrect sentence or start word selections, we instead globally normalize the scores from each stage of our procedure. In a globally normalized model, we define
\begin{equation}
    \mbox{score}(a, d, q) = \phi_\mathrm{sent}(d_i) + \phi_\mathrm{sw}(d_{i,j}) + \phi_\mathrm{ew}(d_{i,j:k}).
\end{equation}
Then, we model
\begin{equation}
    \mathbb{P}(a \mid d, q) = \frac{\exp(\mbox{score}(a, d, q))}{Z},
\end{equation}
where $Z$ is the partition function
\begin{equation}
    Z = \sum_{a' \in \mathcal{A}(d)} \exp(\mbox{score}(a', d, q)).
\end{equation}
In contrast to locally-normalized models, the model is normalized over all possible search paths instead of normalizing each step of search procedure. At inference time, the problem is to find
\begin{equation}
    \arg\max_{a \in \mathcal{A}(d)}  \mathbb{P}(a \mid d, q),
\end{equation}
which can be approximately computed using beam search.

\subsection{Objective and Training}
We minimize the negative log-likelihood on the training set using stochastic gradient descent. For a single example $(a, d, q)$, the negative log-likelihood
\begin{equation}
   -\mbox{score}(a, d, q) + \log Z
\end{equation}
requires an expensive summation to compute $\log Z$. Instead, to ensure learning is efficient, we use beam search during training and early updates \cite{andor2016globally, zhou2015neural, collins2004incremental}. Concretely, we approximate $Z$ by summing only over candidates on the final beam $\mathcal{B}$:
\begin{equation}
    Z \approx \sum_{a' \in \mathcal{B}} \exp(\mbox{score}(a', d, q)).
\end{equation}

At training time, if the gold sequence falls off the beam at step $t$ during decoding, a stochastic gradient step is performed on the partial objective computed through step $t$ and normalized over the beam at step $t$.

\subsection{Implementation}

Our best performing model uses a stack of 3 Bi-LSTMs for the question and document encodings, and a single Bi-LSTM for the end of span prediction. The hidden dimension of all recurrent layers is 200.
I
We use the 300 dimensional 8.4B token Common Crawl GloVe vectors  \cite{pennington2014glove}. Words missing from the Common Crawl vocabulary are set to zero. In our experiments, all architectures considered have sufficient capacity to overfit the training set. We regularize the models by fixing the word embeddings throughout training, dropping out the inputs of the Bi-LSTMs with probability 0.3 and the inputs to the fully-connected layers with probability 0.4 \cite{srivastava2014dropout}, and adding gaussian noise to the recurrent weights with $\sigma=10^{-6}$. Our models are trained using Adam with a learning rate of 0.0005, $\beta_1=0.9$, $\beta_2=0.999$, $\epsilon=10^{-8}$ and a batch size of 32 \cite{kingma2014adam}. 

All our experiments are implemented in Tensorflow \cite{abadi2016tensorflow}, and we tokenize using Ciseau  \cite{RaimanCiseau2017}. Despite performing beam-search during training, our model trains to convergence in under 4 hours through the use of efficient LSTM primitives in CuDNN \cite{chetlur2014cudnn} and batching our computation over examples and search beams. We release our code and augmented dataset.\footnote{\url{https://github.com/baidu-research/GloballyNormalizedReader}}

Our implementation of the GNR is 24.7 times faster at inference time than the official Bi-Directional Attention Flow implementation\footnote{\url{https://github.com/allenai/bi-att-flow}}. Specifically, on a machine running Ubuntu 14 with 40 Intel Xeon 2.6Ghz processors, 386GB of RAM, and a 12GB TitanX-Maxwell GPU, the GNR with beam size 32 and batch size 32 requires $51.58 \pm 0.266$ seconds (mean $\pm$ std)\footnote{All numbers are averaged over 5 runs.} to process the SQUAD validation set. By contrast, the Bi-Directional Attention Flow model with batch size 32 requires $1260.23 \pm 17.26$ seconds. We attribute this speedup to avoiding expensive bi-directional attention mechanisms and making computation conditional on the search beams.

\section{Type Swaps}

\newcommand\replace[2]{\raisebox{-\baselineskip}{\shortstack{\underline{#1}\\{\tiny #2}}}}

\begin{figure}[ht]
\fbox{\parbox{19.5em}{
{\bf Question}: Who said in \replace{April 25, 2011}{December 2012} that the fight would change from military to law enforcement?\\
{\bf Answer}: \replace{Sheryl Sandberg}{Jeh Johnson}\\
{\bf Document (snippet)}:
\dots Basic objectives of the \replace{Cabinet of Japan}{Bush administration} ``war on terror", such as targeting al Qaeda and building international counterterrorism alliances, remain in place. In \replace{April 25, 2011}{December 2012}, \replace{Sheryl Sandberg}{Jeh Johnson}, the General Counsel of the \replace{ministry of education}{Department of Defense}, stated that the military fight will be replaced by a law enforcement operation when speaking at \replace{Ain Shams University}{Oxford University}\dots
}}
\caption{Type Swaps example. Replacements underlined with originals underneath.}
\label{fig:typeswaps}
\end{figure}

In extractive question answering, the set of possible answer spans can be pruned by only keeping answers whose nature (person, object, place, date, etc.) agrees with the question type (Who, What, Where, When, etc.). While this heuristic helps human readers filter out irrelevant parts of a document when searching for information, no explicit supervision of this kind is present in the dataset. Despite this absence, the distribution question representations learned by our models appear to utilize this heuristic. The final hidden state of the question-encoding LSTMs naturally cluster based on question type (Table \ref{table:qtype}).

\begin{table*}[t]
\caption{Top bigrams in K-means ($K=7$) clusters of question after Bi-LSTM. We observe emergent clustering according to question type: e.g. {\em Where}$\to$ Cluster 7, {\em Who}$\to$ Cluster 3. ``What" granularity only observable with more clusters.}
\begin{center}
\begin{tabular}{ |r|r|r|r|r|r|r|r|}
\hline
     Cluster & 1 & 2 & 3 & 4 & 5 & 6 & 7\\
     Size & 84789 & 42187 & 53061 & 130022 & 27549 & 16894 & 28377\\
\hline
Bigram & \multicolumn{7}{|c|}{Bigram Occurences}\\
\hline
  \em what is &  3339 &   520 &    87 &   3736 &    20 &     8 &   138\\
 \em what did &  2463 &     3 &     3 &    112 &     1 &     0 &     1\\
 \em how many &     2 &  5095 &     1 &      1 &     0 &     0 &     0\\
 \em how much &     7 &  1102 &     0 &     12 &     0 &     0 &     0\\
  \em who was &     2 &     0 &  1934 &      0 &     0 &     0 &     1\\
  \em who did &     2 &     0 &   683 &      2 &     0 &     0 &     0\\
 \em what was &  2177 &   508 &   105 &   2034 &    71 &    31 &    92\\
 \em when did &     0 &     0 &     0 &      1 &  2772 &     0 &     0\\
 \em when was &     0 &     0 &     1 &      1 &  1876 &     0 &     0\\
\em what year &     0 &     0 &     0 &      1 &    13 &  2690 &     0\\
  \em in what &    52 &     3 &     9 &    727 &   110 &  1827 &   518\\
\em where did &     0 &     0 &     0 &     13 &     1 &     0 &   955\\
 \em where is &     0 &     1 &     0 &     11 &     0 &     0 &   665\\
 \hline
\end{tabular}
\end{center}
\label{table:qtype}
\end{table*}

In other words, the task induces a question encoding that superficially respects type information. This property is a double-edged sword: it allows the model to easily weed out answers that are inapplicable, but also leads it astray by selecting a text span that shares the answer's type but has the wrong underlying entity. A similar observation was made in the error analysis of \cite{weissenborn2017fastqa}. We propose Type Swaps, an augmentation strategy that leverages this emergent behavior in order to improve the model's ability to prune wrong answers, and make it more robust to surface form variation. This strategy has three steps:
\begin{enumerate}
    \item Locate named entities in document and question.
    \item Collect surface variation for each entity type:
    \subitem \emph{human} $\to$ \{Ada Lovelace, Daniel Kahnemann,...\},
    \subitem \emph{country} $\to$ \{USA, France, ...\}, ...
    \item Generate new document-question-answer examples by swapping each named entity in an original triplet with a surface variant that shares the same type from the collection.
\end{enumerate}

\begin{figure}[ht]
    \centering
    \includegraphics[width=3in]{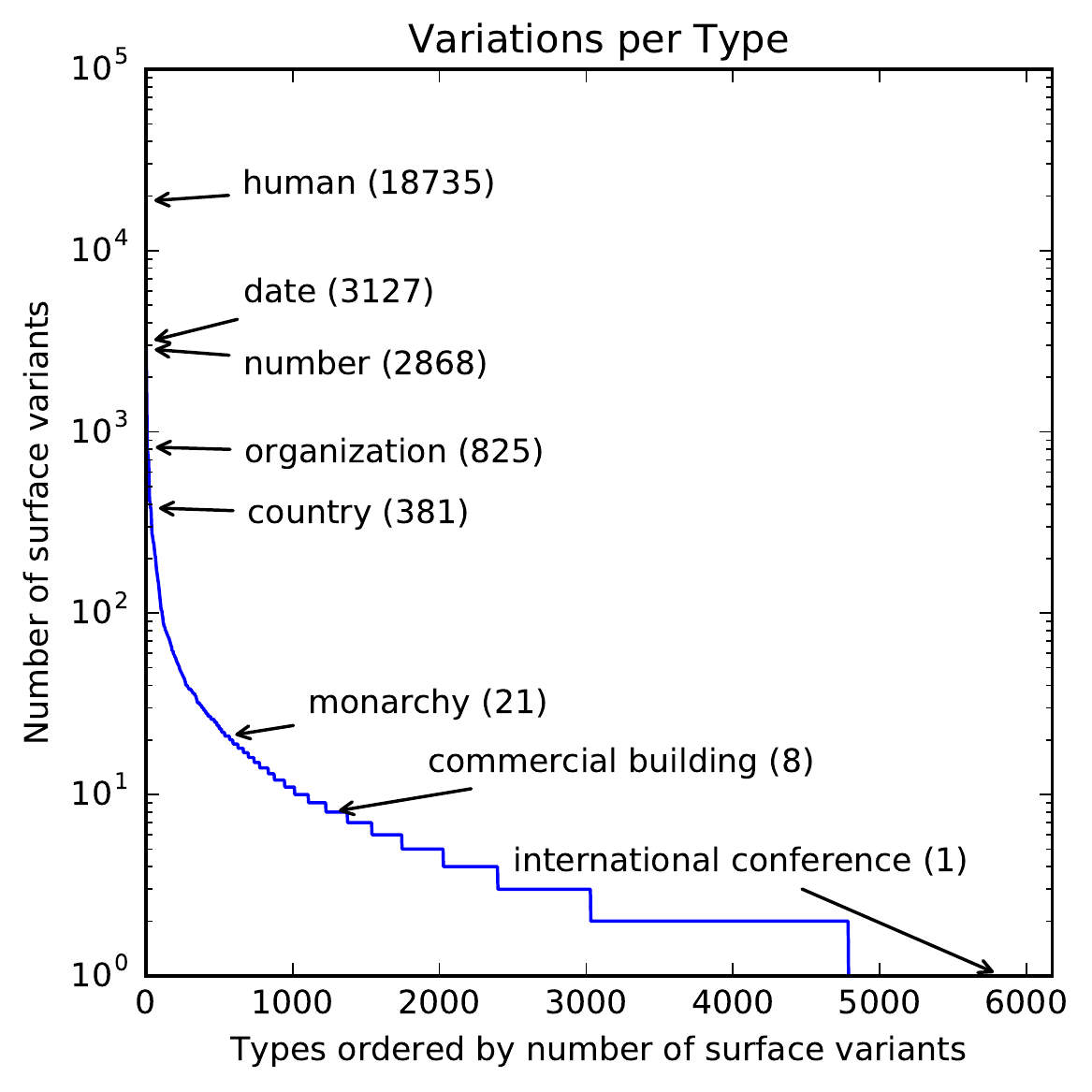}
    \caption{The majority of the surface variations occur for people, numbers, dates, and organizations.}
    \label{fig:variations-type}
\end{figure}

Assigning types to named entities in natural language is an open problem, nonetheless when faced with documents where we can safely assume that the majority of the entities will be contained in a large knowledge base (KB) such as Wikidata \citet{vrandevcic2014wikidata} we find that simple string matching techniques are sufficiently accurate. Specifically, we use a part of speech tagger \cite{spacy} to extract nominal groups in the training data and string-match them with entities in Wikidata. Using this technique, we are able to extract 47,598 entities in SQuAD that fall under 6,380 Wikidata {\tt instance of}\footnote{\url{https://www.wikidata.org/wiki/Property:P31}} types. Additionally we assign ``number types" (e.g. \emph{year}, \emph{day of the week}, \emph{distance}, etc.) to nominal groups that contain dates, numbers, or quantities\footnote{In our experiments we found that not including numerical variation in the generated examples led to an imbalanced dataset and lower final performance.}. These extraction steps produce 84,632 unique surface variants (on average 16.93 per type) with the majority of the variation found in humans, numbers or organizations as visible in Figure \ref{fig:variations-type}. 

With this method, we can generate $2.92 \cdot 10^{369}$  unique documents (average of $3.36 \cdot 10^{364}$ new documents for each original document). To ensure there is sufficient variation in the generated documents, we sample from this set and only keep variations where the question or answer is mutated. At each training epoch, we train on $T$ Type Swap examples and the full original training data. An example output of the method is shown in Figure \ref{fig:typeswaps}.

\section{Results}

\begin{table*}[t]
\caption{Model comparison}
\begin{center}
\begin{tabular}{|l|r|r|}
 \hline
 Model & EM & F1  \\
 \hline
 Human \cite{rajpurkar2016squad} & 80.3 & 90.5\\
 \hline
 {\it Single model} & & \\
 Sliding Window \cite{rajpurkar2016squad} & 13.3 & 20.2 \\
 Match-LSTM \cite{wang2016machine} & 64.1 & 73.9 \\
 DCN \cite{xiong2016dynamic} & 65.4 & 75.6 \\
 Rasor \cite{lee2016learning} & 66.4 & 74.9 \\
 Bi-Attention Flow \cite{seo2016bidirectional} & 67.7 & 77.3 \\
 R-Net\cite{rnet} & {\bf 72.3} & {\bf 80.6}\\
  Globally Normalized Reader w/o Type Swaps (Ours) & 66.6 & 75.0\\
 Globally Normalized Reader (Ours) & 68.4 & 76.21\\
 \hline
\end{tabular}
\end{center}
\label{table:results}
\end{table*}

We evaluate our model on the 100,000 example SQuAD dataset \cite{rajpurkar2016squad} and perform several ablations to evaluate the relative importance of the proposed methods. 

\subsection{Learning to Search}

In our first experiment, we aim to quantify the importance of global normalization on the learning and search process. We use $T=10^4$ Type Swap samples and vary beam width $B$ between 1 and 32 for a locally and globally normalized models and summarize the Exact-Match and F1 score of the model's predicted answer and ground truth computed using the  evaluation scripts from \cite{rajpurkar2016squad} (Table \ref{table:beam}). We additionally report another metric, the {\em Sentence} score, which is a measure for how often the predicted answer came from the correct sentence. This metric provides a measure for where mistakes are made during prediction.

\begin{table}[t]
\caption{Impact of Beam Width $B$}
\begin{center}
\begin{tabular}{|l|r|r|r|r|}
 \hline
 Model & $B$ & EM & F1 & Sentence \\
 \hline
 \multirow{4}{*}{Local, $T=10^4$} & 1  & 65.7 & 74.8 & \bf 89.0 \\
  & 2  & 66.6 & 75.0 & 88.3 \\
  & 10 & 66.7 & 75.0 & 88.6 \\
  & 32 & 66.3 & 74.6 & 88.0 \\
  & 64 & 66.6 & 75.0 & 88.8 \\
 \hline
 \multirow{4}{*}{Global, $T=10^4$} & 1  & 58.8      & 68.4      & 84.5 \\
    & 2  & 64.3      & 73.0      & 86.8 \\
    & 10 & 66.6      & 75.2      & 88.1 \\
    & 32 & \bf 68.4  & \bf 76.21 & 88.4 \\
  & 64 & 67.0 & 75.6 & 88.4 \\
 \hline
\end{tabular}
\end{center}
\label{table:beam}
\end{table}

\subsection{Type Swaps}

In our second experiment, we evaluate the impact of the amount of augmented data on the performance of our model. In this experiment, we use the best beam sizes for each model ($B=10$ for local and $B=32$ for global) and vary the augmentation from $T=0$ (no augmentation) to $T=5\cdot10^4$. The results of this experiment are summarized in (Table \ref{table:augment}).

We observe that both models improve in performance with $T > 0$ and performance degrades past $T=10^4$. Moreover, data augmentation and global normalization are complementary. Combined, we obtain 1.6 EM and 2.0 F1 improvement over the locally normalized baseline.

We also verify that the effects of Type Swaps are not limited to our specific model by observing the impact of augmented data on the DCN+ \cite{xiong2016dynamic}\footnote{
The DCN+ is the DCN with additional hyperparameter tuning by the same authors as submitted on the SQuAD leaderboard \url{https://rajpurkar.github.io/SQuAD-explorer/}.}. We find that it strongly reduces generalization error, and helps improve F1, with potential further improvements coming by reducing other forms of regularization (Table \ref{table:dcnaugment}).

\begin{table}[t]
\caption{Impact of Augmentation Sample Size $T$.}
\begin{center}
\begin{tabular}{|l|r|r|r|r|}
 \hline
 Model & $T$ & EM & F1 & Sentence \\
 \hline
 Local & 0              & 65.8     & 74.0     & 88.0 \\
 Local & $10^3$         & 66.3     & 74.6     & 88.9 \\
 Local & $10^4$         & 66.7     & 74.9     & \bf 89.0 \\
 Local & $5 \cdot 10^4$ & 66.7     & 75.0     & \bf 89.0 \\
 Local & $10^5$         & 66.2     & 74.5     & 88.6 \\
 \hline
 Global & 0              & 66.6     & 75.0     & 88.2 \\
 Global & $10^3$         & 66.9     & 75.0     & 88.1 \\
 Global & $10^4$         & \bf 68.4 & \bf 76.21 & 88.4 \\
 Global & $5 \cdot 10^4$ & 66.8     & 75.3     & 88.3 \\
 Global & $10^5$         & 66.1     & 74.3     & 86.9 \\
 \hline
\end{tabular}
\end{center}
\label{table:augment}
\end{table}

\begin{table}[t]
\caption{Impact of Type Swaps on the DCN+}
\begin{center}
\begin{tabular}{|l|r|r|r|r|}
 \hline
 $T$ & Train F1 & Dev F1 \\
 \hline
 0 & 81.3 & 78.1 \\
 $5\cdot10^4$ & 72.5 & {\bf 78.2} \\
 \hline
\end{tabular}
\end{center}
\label{table:dcnaugment}
\end{table}

\section{Discussion}

In this section we will discuss the results presented in Section 4, and explain how they relate to our main claims.

\subsection{Extractive Question Answering as a Search Problem}
Sentences provide a natural and powerful document decomposition for search that can be easily learnt as a search step: for all the models and configurations considered, the {\em Sentence} score was above 88\% correct (Table 
\ref{table:beam})\footnote{The objective function difference explains the lower performance of globally versus locally normalized models on the {\em Sentence} score: local models must always assign the highest probability to the correct sentence, while global models only ensure the correct span has the highest probability. Thus global models do not need to enforce a high margin between the correct answer's sentence score and others and are more likely to keep alternate sentences around.}. Thus, sentence selection is the easy part of the problem, and the model can allocate more computation (such as the end-word selection Bi-LSTM) to spans likely to contain the answer. This approach avoids wasteful work on unpromising spans and is important for further scaling these methods to long documents.
\subsection{Global Normalization}
The Globally Normalized Reader outperforms previous approaches and achieves the second highest EM behind \cite{rnet}, without using bi-directional attention and only scoring spans in its final beam. Increasing the beam width improves the results for both locally and globally normalized models (Table \ref{table:beam}), suggesting search errors account for a significant portion of the performance difference between models. Models such as \citet{lee2016learning} and \citet{wang2016machine} overcome this difficulty by ranking all possible spans and thus never skipping a possible answer. Even with large beam sizes, the locally normalized model underperforms these approaches. However, by increasing model flexibility and performing search during training, the globally normalized model is able to recover from search errors and achieve much of the benefits of scoring all possible spans.

\subsection{Type-Aware Data Augmentation}

Type Swaps, our data augmentation strategy, offers a way to incorporate the nature of the question and the types of named entities in the answers into the learning process of our model and reduce sensitivity to surface variation. Existing neural-network approaches to extractive QA have so far ignored this information. Augmenting the dataset with additional type-sensitive synthetic examples improves performance by providing better coverage of different answer types. Growing the number of augmented samples used improves the performance of all models under study (Table \ref{table:augment}-\ref{table:dcnaugment}). With $T \in [10^4, 5\cdot 10^4]$, (EM, F1) improve from $(65.8\to66.7, 74.0\to75.0)$ for locally normalized models, and $(66.6\to68.4,75.0\to76.21)$ for globally normalized models. 

Past a certain amount of augmentation, we observe performance degradation. This suggests that despite efforts to closely mimic the original training set, there is a train-test mismatch or excess duplication in the generated examples.

Our experiments are conducted on two vastly different architectures and thus these benefits are expected to carry over to different models \cite{weissenborn2017fastqa,seo2016bidirectional,rnet}, and perhaps more broadly in other natural language tasks that contain named entities and have limited supervised data.

\section{Related Work}

Our work is closely related to existing approaches in learning to search, extractive question answering, and data augmentation for NLP tasks.

\paragraph{Learning to Search.}
Several approaches to learning to search have been proposed for various NLP tasks and conditional computation. Most recently, \citet{andor2016globally} and \citet{zhou2015neural} demonstrated the effectiveness of globally normalized networks and training with beam search for part of speech tagging and transition-based dependency parsing, while \citet{wiseman2016sequence} showed that these techniques could also be applied to sequence-to-sequence models in several application areas including machine translation. These works focus on parsing and sequence prediction tasks and have a fixed computation regardless of the search path, while we show that the same techniques can also be straightforwardly applied to question answering and extended to allow for conditional computation based on the search path.

Learning to search has also been used in context of modular neural networks with conditional computation in the work of \citet{andreas2016learning} for image captioning. In their work reinforcement learning was used to learn how to turn on and off computation, while we find that conditional computation can be easily learnt with maximum likelihood and the help of early updates \cite{andor2016globally, zhou2015neural, collins2004incremental} to guide the training process.

Our framework for conditional computation whereby the search space is pruned by a sequence of increasingly complex models is broadly reminiscent of the structured prediction cascades of \cite{weiss2010structured}. \citet{trischler2016parallel} also explored this approach in the context of question answering.

\paragraph{Extractive Question Answering.}
Since the introduction of the SQuAD dataset, numerous systems have achieved strong results. \citet{seo2016bidirectional,rnet} and \citet{xiong2016dynamic} make use of a bi-directional attention mechanisms, whereas the GNR is more lightweight and achieves similar results without this type of attention mechanism. The document representation used by the GNR is very similar to \citet{lee2016learning}. However, both \citet{lee2016learning} and \citet{wang2016machine} must score all $O(N^2)$ possible answer spans, making training and inference expensive. The GNR avoids this complexity by learning to search during training and outperforms both systems while scoring only $O(B)$ spans. \citet{weissenborn2017fastqa} is a locally normalized model that first predicts start and then end words of each span. Our experiments lead us to believe that further factorizing the problem and using global normalization along with our data augmentation would yield corresponding improvements.

\paragraph{Data augmentation.}
Several works use data augmentation to control the generalization error of deep learning models. \citet{zhang2015text} use a thesaurus to generate new training examples based on synonyms. \citet{vijayaraghavan2016deepstance} employs a similar method, but uses Word2vec and cosine similarity to find similar words. \citet{jia2016data} use a high-precision synchronous context-free grammar to generate new semantic parsing examples. Our data augmentation technique, Type Swaps, is unique in that it leverages an external knowledge-base to provide new examples that have more variation and finer-grained changes than methods that use only a thesaurus or Word2Vec, while also keeping the narrative and grammatical structure intact.

More recently \citet{zhou2017neural} proposed a sequence-to-sequence model to generate diverse and realistic training  question-answer pairs on SQuAD. Similar to their approach, our technique makes use of existing examples to produce new examples that are fluent, however we also are able to explicitly incorporate entity type information into the generation process and use the generated data to improve the performance of question answering models.

\section{Conclusions and Future Work}
In this work, we provide a methodology that overcomes several limitations of existing approaches to extractive question answering. In particular, our proposed model, the Globally Normalized Reader, reduces the computational complexity of previous models by casting the question answering as search and allocating more computation to promising answer spans. Empirically, we find that this approach, combined with global normalization and beam search during training, leads to near state of the art results. Furthermore, we find that a type-aware data augmentation strategy improves the performance of all models under study on the SQuAD dataset. The method is general, only requiring that the training data contains named entities from a large KB. We expect it to be applicable to other NLP tasks that would benefit from more training data.

As future work we plan to apply  the GNR to other question answering datasets such as MS MARCO \cite{nguyen2016ms} or NewsQA \cite{trischler2016newsqa}, as well as investigate the applicability and benefits of Type Swaps to other tasks like named entity recognition, entity linking, machine translation, and summarization. Finally, we believe there a broad range of structured prediction problems (code generation, generative models for images, audio, or videos) where the size of original search space makes current techniques intractable, but if cast as learning-to-search problems with conditional computation, might be within reach.

\section*{Acknowledgments}
We would like to thank the anonymous reviewers for their valuable feedback. In addition, we thank Adam~Coates, Carl~Case, Andrew~Gibiansky, and Szymon~Sidor for thoughtful comments and fruitful discussion. We also thank James~Bradbury and Bryan~McCann for running Type Swap experiments on the DCN+.

\bibliography{emnlp2017}
\bibliographystyle{emnlp_natbib}

\end{document}